\documentclass[10pt,twocolumn,letterpaper]{article}

\usepackage{iccv}
\usepackage{times}
\usepackage{epsfig}
\usepackage{graphicx}
\usepackage{amsmath}
\usepackage{amssymb}

\usepackage{amsfonts}
\usepackage{mathtools}
\usepackage{bm}
\usepackage{xfrac}
\usepackage{xspace}
\usepackage[makeroom]{cancel}
\usepackage{makecell}
\usepackage{multirow}
\usepackage{subcaption}
\usepackage{color}
\usepackage{xcolor}
\usepackage{colortbl}
\usepackage{lipsum}
\usepackage{ wasysym }
\usepackage{textcomp}
\usepackage[abs]{overpic}

\usepackage{booktabs}
\definecolor{myyellow}{rgb}{1,1, 0.6}
\definecolor{myorange}{rgb}{1, 0.8, 0.6}
\definecolor{myred}{rgb}{1, 0.6, 0.6}
\definecolor{blue-violet}{rgb}{0.54, 0.17, 0.89}

\usepackage[pagebackref=true,breaklinks=true,letterpaper=true,colorlinks,bookmarks=false]{hyperref}

\iccvfinalcopy 

\ificcvfinal\fi

\begin{document}

\title{HOSNeRF: Dynamic Human-Object-Scene Neural Radiance Fields from a Single Video}

\author{%
  Jia-Wei Liu$\textsuperscript{\rm 1}$\thanks{Work is partially done during internship at ARC Lab, Tencent PCG.},  Yan-Pei Cao$\textsuperscript{\rm 2}$, Tianyuan Yang$\textsuperscript{\rm 1}$, Eric Zhongcong Xu$\textsuperscript{\rm 1}$, Jussi Keppo$\textsuperscript{\rm 4,5}$, \\ 
  Ying Shan$\textsuperscript{\rm 2}$, Xiaohu Qie$\textsuperscript{\rm 3}$, Mike Zheng Shou$\textsuperscript{\rm 1}$\thanks{Corresponding Author.} \\ 
  \\
  $~\textsuperscript{\rm 1}$ Show Lab, National University of Singapore $~\textsuperscript{\rm 2}$ ARC Lab, $~\textsuperscript{\rm 3}$ Tencent PCG\\
  $~\textsuperscript{\rm 4}$ Business School, $~\textsuperscript{\rm 5}$ Institute of Operations Research and Analytics, National University of Singapore
}

\maketitle
\ificcvfinal\thispagestyle{empty}\fi

\newcommand*{\method}{HOSNeRF}

\begin{abstract}

We introduce \method{}, a novel 360{\textdegree} free-viewpoint rendering method that reconstructs neural radiance fields for dynamic human-object-scene from a single monocular in-the-wild video. Our method enables pausing the video at any frame and rendering all scene details (dynamic humans, objects, and backgrounds) from arbitrary viewpoints. The first challenge in this task is the complex object motions in human-object interactions, which we tackle by introducing the new object bones into the conventional human skeleton hierarchy to effectively estimate large object deformations in our dynamic human-object model. The second challenge is that humans interact with different objects at different times, for which we introduce two new learnable object state embeddings that can be used as conditions for learning our human-object representation and scene representation, respectively. Extensive experiments show that \method{} significantly outperforms SOTA approaches on two challenging datasets by a large margin of $40\%\sim50\%$ in terms of LPIPS. The code, data, and compelling examples of 360{\textdegree} free-viewpoint renderings from single videos will be released in \url{https://showlab.github.io/HOSNeRF}.

\end{abstract}

\section{Introduction}

\begin{figure*}[h]
\begin{centering}
\includegraphics[width=1\linewidth]{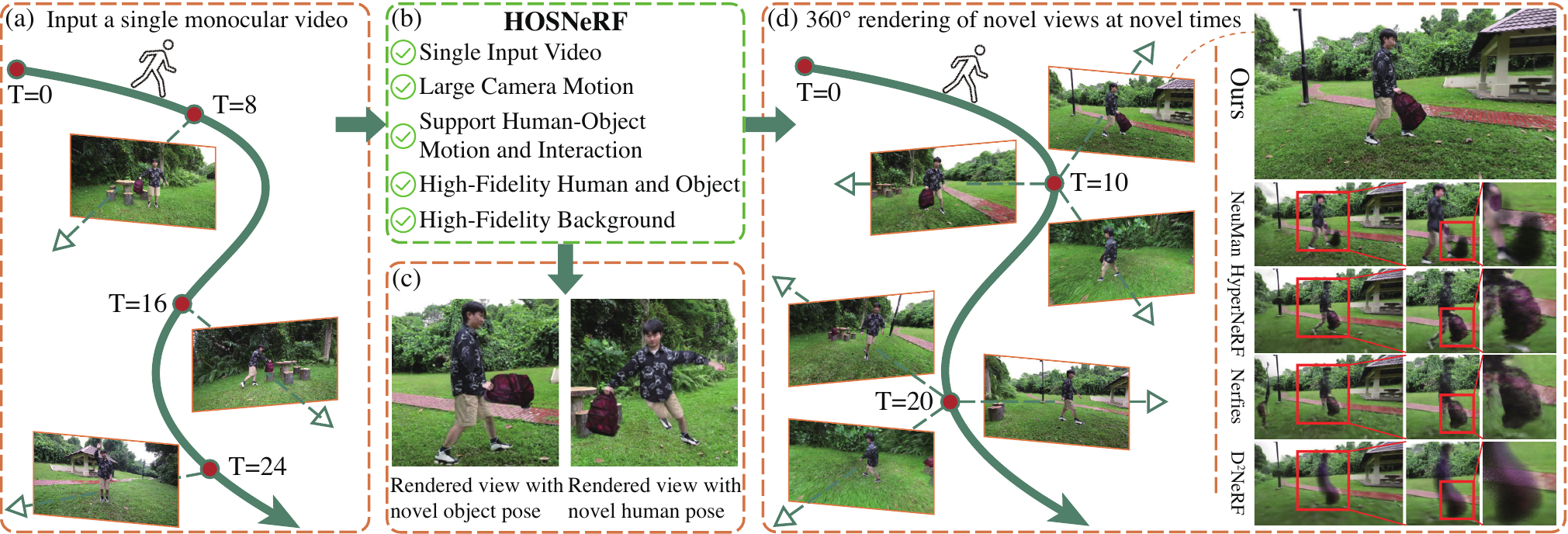}
\par\end{centering}
\vspace{-2mm}
\caption{\label{fig:teaser}Our \method{} \textbf{(b)} takes a single monocular in-the-wild video \textbf{(a)} as input, and creates high-fidelity 360{\textdegree} free-viewpoint rendering of all scene details (dynamic human body, objects, and background) at any time \textbf{(d)}. Our method enables rendering views with novel object poses and novel human poses as shown in \textbf{(c)}, and produces high-fidelity dynamic novel view synthesis results at novel timesteps, with significant improvements over SOTA approaches as shown in \textbf{(d)}.}
\vspace{-5mm}
\end{figure*}

Video reconstruction and free-viewpoint rendering offer innovative opportunities for creating immersive experiences, encompassing virtual reality, telepresence, metaverse, and 3D animation production. While reconstructing videos has the potential to enhance user engagement and provide more realistic environments, it also poses significant challenges in terms of monocular viewpoints and complicated human-environment interactions.

In recent years, remarkable progress has been made in novel view synthesis, particularly since the introduction of Neural Radiance Fields (NeRF)~\cite{mildenhall2021nerf}. While initially limited to reconstructing static 3D scenes based on multi-view images, subsequent studies have proposed various approaches to address the challenge of dynamic view synthesis. NeRF-based techniques have evolved to either incorporate deformation fields that map dynamic fields to canonical NeRF spaces~\cite{pumarola2021d,park2021nerfies,park2021hypernerf,tretschk2021non}, or model dynamic scenes as 4D spatio-temporal radiance fields~\cite{li2021neural,gao2021dynamic}. While these approaches have shown promising results in dynamic view synthesis, they are limited to simple deformations. Another line of research is specifically designed for dynamic neural human modeling that relies on estimated human poses as a priori information~\cite{peng2021neural,weng2022humannerf}. Recently, Neuman~\cite{jiang2022neuman} combines pose-driven dynamic human models with static scene models for representing dynamic human-centric scenes. 

However, none of the aforementioned techniques can accurately reconstruct challenging monocular videos with fast and complex human-object-scene motions and interactions, as shown in Fig.~\ref{fig:teaser}(d). This is due to two particular challenges listed below. To tackle them, we introduce a novel method called \textbf{H}uman-\textbf{O}bject-\textbf{S}cene \textbf{Ne}ural \textbf{R}adiance \textbf{F}ields (\method{}).

\textbf{i}) \textbf{Complex object motions in human-object interactions.} In contrast to the simple motions that can be modeled by general deformation modules~\cite{park2021nerfies,park2021hypernerf}, the object motion during human-object interaction is often drastic and composed of various atomic motions (\emph{e.g.}, play tennis). To tackle this challenge, we propose new object bones that are attached to the human skeleton hierarchy to estimate human-object deformations in a coarse-to-fine manner for our dynamic human-object model. The object bones and underlying object linear blend skinning (object LBS) allow for the accurate estimation of objects' deformations through the relative transformations in the kinematic tree of the skeleton hierarchy. 

\textbf{ii}) \textbf{Humans interact with different objects at different times.} The above human-object model is designed for humans interacting with the same object over time. But when the person puts down the current object or picks a new object, it is not clear how to dynamically remove/add such objects in the static background model and the human-object model whose canonical space is static. To handle this challenge, we introduce two new learnable object state embeddings that can be used as conditions for learning our human-object representation and scene representation, respectively.

Finally, we systematically explore and identify effective training objectives and strategies for our proposed \method{}, including deformation cycle consistency, optical flow supervisions, and foreground-background rendering. On two challenging datasets collected by ourselves and NeuMan~\cite{jiang2022neuman}, our \method{} achieves high-fidelity dynamic novel view synthesis results and enables pausing the monocular video at any time and rendering all scene details (dynamic humans, objects, and backgrounds) from arbitrary viewpoints, as shown in Fig.~\ref{fig:teaser}(d). 

In summary, our main contributions are:

\vspace{-2mm}
\begin{itemize}
\item    
We present a novel framework of \method{}, the first work to achieve 360{\textdegree} free-viewpoint high-fidelity novel view synthesis for dynamic scenes with human-environment interactions from a single video.
\item
We propose the object bones and state-conditional representations to handle the non-rigid motions and interactions of humans, objects, and the environment. 
\item 
Extensive experiments show that \method{} significantly outperforms SOTA approaches on two challenging datasets by $40\%\sim50\%$ in terms of LPIPS.
\end{itemize}

\section{Related Work}

\subsection{Dynamic Human Modeling}

Dynamic human modeling has shown promising results in utilizing various representations such as point clouds~\cite{aliev2020neural, wu2020multi}, meshes~\cite{habermann2021real, liu2019neural}, voxels~\cite{lombardi2019neural, sitzmann2019deepvoxels}, and neural implicit functions~\cite{huang2020arch, saito2020pifuhd}, with models like SMPL~\cite{loper2015smpl, prokudin2021smplpix} being commonly used for parameterizing the human body. Since the introduction of NeRF~\cite{mildenhall2021nerf}, neural human representation~\cite{peng2021neural,shao2022doublefield,su2021nerf,xu2021h,liu2021neural} has achieved remarkable progress on representing dynamic human bodies from sparse-view videos. Among them, Neural Actor~\cite{liu2021neural} and Neural Body~\cite{peng2021neural} pioneer in combining NeRF~\cite{mildenhall2021nerf} with SMPL deformable meshes to represent human bodies with complex motions. Subsequent works have further improved on the generalizability~\cite{kwon2021neural,chen2022geometry,9888037} and animatability~\cite{peng2021animatable,li2022tava} of human bodies. To support multi-person modeling, recent works~\cite{zhang2021editable,shuai2022novel} have proposed to segment each human into 3D bounding boxes and learn a separate layered dynamic NeRF for each person. Other works~\cite{suo2021neuralhumanfvv, su2022robustfusion, jiang2022neuralhofusion, sun2021neural} are specifically designed to reconstruct the dynamic human and object with RGB-D or multi-view videos as inputs. They track the human and object pose, and separately reconstruct them with volumetric fusion~\cite{su2022robustfusion}, neural texture blending~\cite{sun2021neural}, or neural rendering~\cite{jiang2022neuralhofusion,zhang2022neuraldome}.

Despite achieving promising results, these approaches require multi-view videos or RGB-D as input, limiting their real-world applications. To solve this problem, HumanNeRF~\cite{weng2022humannerf} is proposed to represent moving humans from a monocular video by the human pose-driven deformation module and canonical space. NeuMan~\cite{jiang2022neuman} is the first successful attempt at reconstructing both the dynamic human and static background from a single video. However, Neuman~\cite{jiang2022neuman} does not support human-environment interactions and performs poorly at large camera motions.

\subsection{Dynamic View Synthesis for General Scenes}

Most prior approaches on dynamic scene modeling require synchronized multi-view videos~\cite{zitnick2004high,li2022neural,stich2008view,collet2015high,zhang2021editable,wang2022fourier} or depth~\cite{zollhofer2014real,innmann2016volumedeform,newcombe2015dynamicfusion} as input. Recent studies have built upon NeRF~\cite{mildenhall2021nerf} to reconstruct dynamic neural radiance fields from monocular videos by either learning a deformation field that maps dynamic observation to canonical field~\cite{pumarola2021d,park2021nerfies,park2021hypernerf,tretschk2021non} or building 4D spatio-temporal radiance fields~\cite{xian2021space,li2021neural,gao2021dynamic}. Among them, Nerfies~\cite{park2021nerfies} associates latent codes with the deformation field and HyperNeRF~\cite{park2021hypernerf} represents motion in a high-dimensional space. D$^2$NeRF~\cite{wu2022d} builds upon HyperNeRF~\cite{park2021hypernerf} and further decouples the dynamic components from the static background, and represents them separately with a HyperNeRF~\cite{park2021hypernerf} and NeRF~\cite{mildenhall2021nerf}. DynIBaR~\cite{li2022dynibar} proposes a motion-adjusted multi-view feature aggregation module to synthesize new viewpoints by aggregating features from nearby views. Other studies have introduced voxel grids~\cite{fang2022fast,liu2022devrf,song2022nerfplayer} or planar representations~\cite{fridovich2023k,cao2023hexplane} for fast dynamic radiance fields reconstruction. While these approaches have achieved high-fidelity dynamic view synthesis results, they are restricted to simple scene deformations. In contrast, our \method{} is capable of representing significant human-object motions and interactions in complex environments.

\section{Method}

\subsection{Preliminaries} \label{sec:humannerf}

\noindent\textbf{HumanNeRF~\cite{weng2022humannerf}} has been recently introduced to represent a moving person with a NeRF~\cite{mildenhall2021nerf}-based canonical space $\Psi_{\mathrm{c}}$ that maps 3D points to color $\mathbf{c}$ and density $d$, and a human pose-guided deformation field $\Psi_{\mathrm{d}}$ that transforms deformed points $\mathbf{x}^{i}_{\mathrm{d}}$ from the deformed space to canonical points $\mathbf{x}^{i}_{\mathrm{c}}$ in the canonical space ($i$ omitted for simplicity).
\begin{equation}
\Psi_{\mathrm{c}}\left(\gamma\left(\mathbf{x}_{\mathrm{c}}\right)\right)\longmapsto\left(\mathbf{c},\,\sigma\right),\;\Psi_{\mathrm{d}}\left(\mathbf{x}_{\mathrm{d}},\,\mathcal{J},\,\mathcal{R}\right)\longmapsto\left(\mathbf{x}_{\mathrm{c}}\right)\,,\label{eq:canonical}
\end{equation}
where $\gamma\left(\mathbf{x}\right)$ is the standard positional encoding function, and $\mathcal{J}=\left\{ \mathbf{J}_{i}\right\} \,$ and $\mathcal{R}=\left\{ \boldsymbol{\omega}_{i}\right\} \,$ represent the 3D human joints and the local joint axis-angle rotations.

The deformation field $\Psi_{\mathrm{d}}$ is decomposed into the coarse human skeleton-driven deformation $\Psi_{\mathrm{d}}^{\mathrm{coarse}}$, and the fine non-rigid deformation conditioned on human poses $\Psi_{\mathrm{d}}^{\mathrm{fine}}$:
\begin{equation}
\mathbf{x}_{\mathrm{c}}^{\prime}=\Psi_{\mathrm{d}}^{\mathrm{coarse}}\left(\mathbf{x}_{\mathrm{d}},\,\mathcal{J},\,\mathcal{R}\right),\;\mathbf{x}_{\mathrm{c}}=\mathbf{x}_{\mathrm{c}}^{\prime}+\Psi_{\mathrm{d}}^{\mathrm{fine}}\left(\mathbf{x}_{\mathrm{c}}^{\prime},\,\mathcal{R}\right)\,.
\end{equation}

\noindent\textbf{Mip-NeRF 360~\cite{barron2022mip}} is designed to synthesize realistic views for highly intricate, unbounded real-world static scenes. To render pixel colors, the casted rays are split into a set of intervals $T_i = [t_i, t_{i+1})$. The mean and covariance of the conical frustum corresponding to every interval are computed as $(\boldsymbol{\mu}, \boldsymbol{\Sigma}) = \mathbf{r}(T_i)$. To parameterize the Gaussian parameters for unbounded scenes, Mip-NeRF 360~\cite{barron2022mip} further proposes a contraction function $f\left(\mathbf{x}\right)$ that distributes distant points proportionally to disparity,
\begin{equation}
f\left(\mathbf{x}\right)=\left\{ \begin{array}{cc}
\mathbf{x} & \left\Vert \mathbf{x}\right\Vert \leq1\\
\left(2-\frac{1}{\left\Vert \mathbf{x}\right\Vert }\right)\left(\frac{\mathbf{x}}{\left\Vert \mathbf{x}\right\Vert }\right) & \left\Vert \mathbf{x}\right\Vert >1
\end{array}\right..\,
\end{equation}
Then, $f\left(\mathbf{x}\right)$ is applied to $(\boldsymbol{\mu}, \boldsymbol{\Sigma})$ as follows:
\begin{equation}
\left(\boldsymbol{\hat{\mu}},\,\boldsymbol{\hat{\Sigma}}\right)=\left(f\left(\boldsymbol{\mu}\right),\,\mathbf{J}_{f}\left(\boldsymbol{\mu}\right)\boldsymbol{\Sigma}\mathbf{J}_{f}\left(\boldsymbol{\mu}\right)^{\mathrm{T}}\right)\,,
\end{equation}
\noindent where $\mathbf{J}_{f}(\boldsymbol{\mu})$ is the Jacobian of $f$ at $\boldsymbol{\mu}$.
The contracted Gaussian parameters $(\boldsymbol{\hat{\mu}}, \boldsymbol{\hat{\Sigma}})$ are further encoded through the integrated positional encoding (IPE)~\cite{barron2022mip}:
\begin{equation}
\resizebox{2.95in}{!}{$
\hat{\gamma}(\boldsymbol{\hat{\mu}}, \boldsymbol{\hat{\Sigma}}) 
= \Bigg\{ \begin{bmatrix} \sin(2^\ell \boldsymbol{\hat{\mu}}) \exp\left(-2^{2\ell - 1} \operatorname{diag}\left(\boldsymbol{\hat{\Sigma}}\right)\right)
\\ \cos(2^\ell \boldsymbol{\hat{\mu}}) \exp\left(-2^{2\ell - 1} \operatorname{diag}\left(\boldsymbol{\hat{\Sigma}}\right)\right)\end{bmatrix} \Bigg\}_{\ell=0}^{L-1}\,,
$}\label{eq:IPE}
\end{equation}
\noindent and the color and density of intervals can be obtained as
\begin{equation}
\Psi_{\mathrm{s}}\left(\hat{\gamma}\left(\boldsymbol{\hat{\mu}},\,\boldsymbol{\hat{\Sigma}}\right)\right)\longmapsto\left(\mathbf{c},\,\sigma\right)\,,
\end{equation}
\noindent where $\Psi_{\mathrm{s}}$ is the scene NeRF MLP~\cite{barron2022mip}.

\subsection{Dynamic Human-Object Model} \label{sec:dynamic-human-obj}

\begin{figure*}
\begin{centering}
\includegraphics[width=1\linewidth]{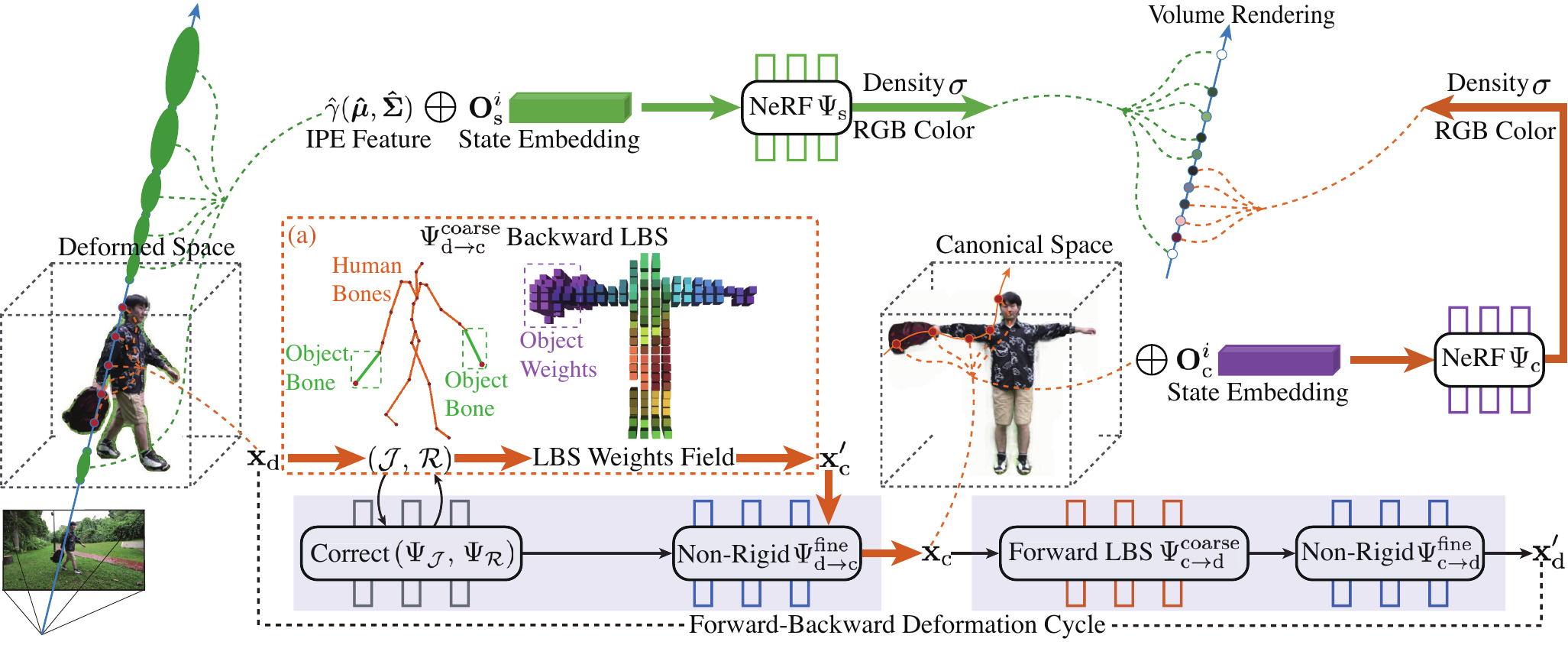}
\par\end{centering}
\vspace{-2mm}
\caption{\label{fig:pipeline}\textbf{Overview of our method.} (1) \textbf{Orange flowchart:} The deformation from deformed points to canonical points are effectively estimated by the human-object backward LBS \textbf{(a)} and non-rigid deformation module, and their properties (\emph{i.e.}, density, color) can be obtained by querying the state-conditional canonical space. (2) \textbf{Green flowchart:} The background intervals sampled from a deformed frame are concatenated with object state embeddings for querying properties through the state-conditional scene model. (3) Novel views can be accordingly synthesized by volume rendering for re-ordered properties.}
\vspace{-5mm}
\end{figure*}

\noindent\textbf{Motivation.} With estimated 3D human poses a priori, HumanNeRF~\cite{weng2022humannerf} is effective at modeling moving people, but can only encode body parts, lacking the capability to model additional structures (\emph{e.g.}, objects held by the person) and hence not suitable for challenging scenarios containing complex human-object interactions. 

A naive approach would be attaching the objects directly to the interacting body parts (\emph{e.g.}, hands) so that the object transformations can be queried via the skeletal motion of the corresponding human joints. However, this approach cannot represent the relative transformations between the object and its interacting body part, making it difficult to model large objects, such as the suitcase shown in Fig.~\ref{fig:Ablation_suitcase}. 

\noindent\textbf{Object Bones.} To address this limitation, based on the conventional human skeleton hierarchy, we introduce a new ``object bone" for each hand, where the starting point of the object bone is connected to the corresponding hand joint. As shown in Fig.~\ref{fig:pipeline} (a), two new object bones (green color) are connected to the hand joints, which represent the relative rotation and translation with respect to their parent hand joints. In potential, the number, bone size, and attaching joints of object bones can also be flexibly customized based on different human-object interaction scenarios. Therefore, our 3D human-object pose consists of 3D joints $\mathcal{J}=\left\{\mathbf{J}_{\mathrm{human}},\,\mathbf{J}_{\mathrm{object}}\right\}$ and their axis-angle rotations $\mathcal{R}=\left\{ \boldsymbol{\omega}_{\mathrm{human}},\,\boldsymbol{\omega}_{\mathrm{object}}\right\}$. 

\noindent\textbf{Object LBS.} To drive the articulated motions of the objects and human body, we learn a linear blend skinning (LBS) weights field during training, as shown in Fig.~\ref{fig:pipeline} (a), which encodes the influence regions of object bones and human bones. Following HumanNeRF~\cite{weng2022humannerf}, we pack the human and object bone blend weights and additional background weight into a single volume with $K+1$ channels, i.e., $\mathcal{W}_{\mathrm{c}\rightarrow\mathrm{d}}\left(\mathbf{x}_{\mathrm{c}}\right)=\left\{w_{\mathrm{c}\rightarrow\mathrm{d}}^{i}\left(\mathbf{x}_{\mathrm{c}}\right)\right\}$, and generate the volume from a learnable latent code using a CNN.

For instance, in Fig.~\ref{fig:pipeline}, the subject holds a backpack in the right hand, so the learned LBS weight field only encodes the influence region of the right object bone with a similar shape to the backpack, while the left object bone has no influence on skeleton motion. Therefore, this learnable LBS weights field enables our object bones to be generalizable to various human-object interaction scenarios with left, right, or both hands. In addition, our human-object skeleton hierarchy enables \method{} to support rendering with novel object pose and novel human pose, as illustrated in Fig.~\ref{fig:teaser}(c).

\noindent\textbf{Forward-Backward Deformation Framework.} To improve the smoothness and consistency of human-object deformations, we propose to leverage the \emph{cycle consistency} between forward and backward human-object deformations. The backward deformation first transforms sampled deformed points $\mathbf{x}_{\mathrm{d}}$ to their corresponding canonical points $\mathbf{x}_{\mathrm{c}}$, which are further mapped back to the deformed space $\mathbf{\hat{x}}_{\mathrm{d}}$ through the forward deformation module.

As illustrated in Fig.~\ref{fig:pipeline}(a), with the proposed human-object skeleton definition and LBS weight fields, the coarse backward skeleton deformation $\Psi_{\mathrm{d}\rightarrow\mathrm{c}}^{\mathrm{coarse}}$ is defined similar to HumanNeRF~\cite{weng2022humannerf},
\begin{equation}
\Psi_{\mathrm{d}\rightarrow\mathrm{c}}^{\mathrm{coarse}}\left(\mathbf{x}_{\mathrm{d}},\,\mathcal{J},\,\mathcal{R}\right)=\sum_{i=1}^{K}w_{\mathrm{d}\rightarrow\mathrm{c}}^{i}\left(\mathbf{x}_{\mathrm{d}}\right)\left(\mathbf{R}_{\mathrm{d}\rightarrow\mathrm{c}}^{i}\mathbf{x}_{\mathrm{d}}+\mathbf{t}_{\mathrm{d}\rightarrow\mathrm{c}}^{i}\right),\,
\end{equation}
\noindent where $K$ is the number of human-object bones, $w_{\mathrm{d}\rightarrow\mathrm{c}}^{i}$ is the blend weight for the $i$-th bone at the deformed space, and $\mathbf{R}_{\mathrm{d}\rightarrow\mathrm{c}}^{i}$ and $\mathbf{t}_{\mathrm{d}\rightarrow\mathrm{c}}^{i}$ are the backward rotation and translation that are explicitly derived from the human-object pose. 

To model the non-rigid motion that can not be represented by skeleton motion, we additionally define the backward fine non-rigid motion $\Psi_{\mathrm{d}\rightarrow\mathrm{c}}^{\mathrm{fine}}$ conditioned on the human-object poses~\cite{weng2022humannerf}:
\begin{equation}
\Delta\mathbf{x}_{\mathrm{d}\rightarrow\mathrm{c}}=\Psi_{\mathrm{d}\rightarrow\mathrm{c}}^{\mathrm{fine}}\left(\gamma\left(\Psi_{\mathrm{d}\rightarrow\mathrm{c}}^{\mathrm{coarse}}\left(\mathbf{x}_{\mathrm{d}},\,\mathcal{J},\,\mathcal{R}\right)\right),\,\mathcal{R}\right)\,.
\end{equation}
Therefore, $\mathbf{x}_{\mathrm{d}}$ can be mapped to the canonical space by
\begin{equation}
\mathbf{x}_{\mathrm{c}}=\Psi_{\mathrm{d}\rightarrow\mathrm{c}}^{\mathrm{coarse}}\left(\mathbf{x}_{\mathrm{d}},\,\mathcal{J},\,\mathcal{R}\right)+\Delta\mathbf{x}_{\mathrm{d}\rightarrow\mathrm{c}}\,.\label{eq:flow_func}
\end{equation}
Accordingly, the forward coarse skeleton motion driven by the human-object pose can be defined as
\begin{equation}
\Psi_{\mathrm{c}\rightarrow\mathrm{d}}^{\mathrm{coarse}}\left(\mathbf{x}_{\mathrm{c}},\,\mathcal{J},\,\mathcal{R}\right)=\sum_{i=1}^{K}w_{\mathrm{c}\rightarrow\mathrm{d}}^{i}\left(\mathbf{x}_{\mathrm{c}}\right)\left(\mathbf{R}_{\mathrm{c}\rightarrow\mathrm{d}}^{i}\mathbf{x}_{\mathrm{c}}+\mathbf{t}_{\mathrm{c}\rightarrow\mathrm{d}}^{i}\right),\,
\end{equation}
\noindent where $w_{\mathrm{c}\rightarrow\mathrm{d}}^{i}$ is the canonical blend weight for the $i$-th bone, $\mathbf{R}_{\mathrm{c}\rightarrow\mathrm{d}}^{i}$ and $\mathbf{t}_{\mathrm{c}\rightarrow\mathrm{d}}^{i}$ are the forward rotation and translation obtained from the human-object pose. In addition, the forward fine non-rigid motion $\Psi_{\mathrm{c}\rightarrow\mathrm{d}}^{\mathrm{fine}}$ is
\begin{equation}
\Delta\mathbf{x}_{\mathrm{c}\rightarrow\mathrm{d}}=\Psi_{\mathrm{c}\rightarrow\mathrm{d}}^{\mathrm{fine}}\left(\gamma\left(\Psi_{\mathrm{c}\rightarrow\mathrm{d}}^{\mathrm{coarse}}\left(\mathbf{x}_{\mathrm{c}},\,\mathcal{J},\,\mathcal{R}\right)\right),\,\mathcal{R}\right)\,.
\end{equation}
\noindent Therefore, the canonical points $\mathbf{x}_{\mathrm{c}}$ can be transformed back to the deformed scene by
\begin{equation}
\mathbf{\hat{x}}_{\mathrm{d}}=\Psi_{\mathrm{c}\rightarrow\mathrm{d}}^{\mathrm{coarse}}\left(\mathbf{x}_{\mathrm{c}},\,\mathcal{J},\,\mathcal{R}\right)+\Delta\mathbf{x}_{\mathrm{c}\rightarrow\mathrm{d}}\,.\label{eq:cycle_func}
\end{equation}

\subsection{State-Conditional Representation}

\begin{figure}
\begin{centering}
\includegraphics[width=1\linewidth]{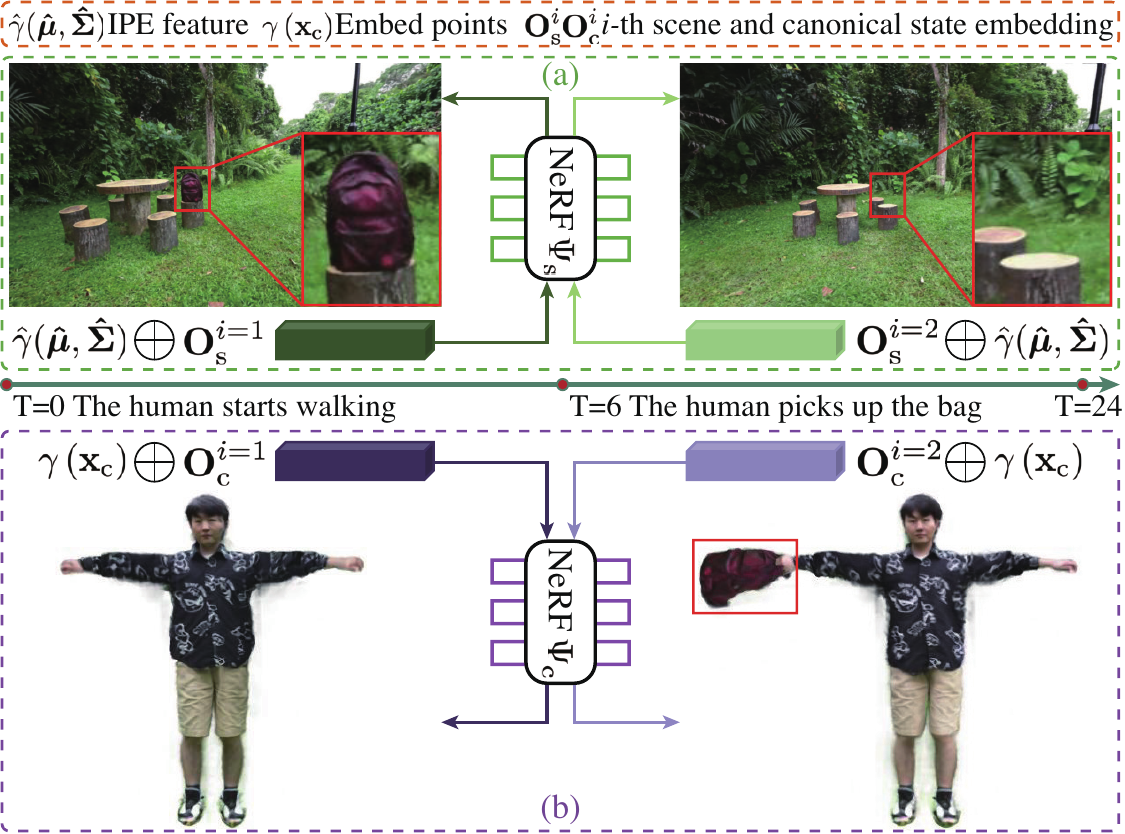}
\par\end{centering}
\vspace{-2mm}
\caption{\label{fig:state}State-conditional representation for the scene model (a) and human-object canonical space (b).}
\vspace{-5mm}
\end{figure}

\noindent\textbf{Motivation.} In complex dynamic scenes, humans can interact with different objects at different timesteps. As the example in Fig.~\ref{fig:state} shows, the human picks up the bag at $T=6$ and puts it down at $T=24$, resulting in $3$ object states: the bag in the background during $T\in\left[0,\,6\right]\,$, the bag held by the human during $T\in\left[6,\,24\right]\,$, the bag at a new position in the background when $T>24\,$. 
Such object state changes prohibit us from directly using the static background model (\emph{i.e.}, Mip-NeRF 360~\cite{barron2022mip}) and the dynamic human-object model proposed in Sec.~\ref{sec:dynamic-human-obj}.

A naive solution is to train a separate network for each object state. However, this solution is limited because each network is trained only on a short segment (\emph{e.g.}, $T\in\left[0,\,6\right]\,$) instead of the whole video.
As a result, some regions often are difficult to reconstruct because they are not observed in the short segment. But these regions could have already been captured by video segments of other object states.

To make full use of all segments to jointly train one model shared for different states, we introduce learnable state embeddings to represent different object states. As depicted in Fig.~\ref{fig:state}, we have two state embeddings respectively for our scene model and human-object model, which are explained in detail as follows.

\noindent\textbf{State-Conditional Scene Model.} As shown in Fig.~\ref{fig:state}(a), we develop a state-conditional Mip-NeRF 360~\cite{barron2022mip} to represent the static scene with temporal object transitions. In a dynamic scene with $N$ object states, we define $N$ learnable state embeddings $\mathcal{O}_{\mathrm{s}}=\left\{ \mathbf{O}_{\mathrm{s}}^{i}\right\} \,\left(i=1,2,\cdots,N\right)$ to represent states information. Therefore, at state $i$, we concatenate the IPE features $\hat{\gamma}(\boldsymbol{\hat{\mu}}, \boldsymbol{\hat{\Sigma}})$ (Eq. (\ref{eq:IPE})) of ray intervals with the state embedding $\mathbf{O}_{\mathrm{s}}^{i}$ as input to the scene MLP $\Psi_{\mathrm{s}}$ for querying the scene color and density.
\begin{equation}
\Psi_{\mathrm{s}}\left(\mathrm{concat}\left(\hat{\gamma}\left(\boldsymbol{\hat{\mu}},\,\boldsymbol{\hat{\Sigma}}\right),\,\mathbf{O}_{\mathrm{s}}^{i}\right)\right)\longmapsto\left(\mathbf{c},\,\sigma\right)\,.
\end{equation}
\noindent This representation enables a shared scene model across the video with multiple object state transitions. 

\noindent\textbf{State-Conditional Dynamic Human-Object Model.} In Sec.~\ref{sec:dynamic-human-obj}, we introduce the human-object pose-driven deformation module and the human-object canonical space for modeling dynamic human-objects. However, the reconstructed canonical space is insufficient in representing the temporal changes in object geometry and appearance when the subject interacts with new objects. To address this limitation, we condition the canonical space on the object states, as shown in Fig.~\ref{fig:state}(b). In particular, we employ an object state-conditional canonical space and optimize a shared blend weights field across all states.

In a dynamic scene with $N$ object states, we define $N$ learnable state embeddings $\mathcal{O}_{\mathrm{c}}=\left\{ \mathbf{O}_{\mathrm{c}}^{i}\right\} \,\left(i=1,2,\cdots,N\right)$ to represent object states in the canonical space. Therefore, at object state $i$, we concatenate the positionally encoded canonical points in Eq. (\ref{eq:canonical}) with the state embedding $\mathbf{O}_{\mathrm{c}}^{i}$ as input to the canonical MLP for querying the human-object color and density.
\begin{equation}
\Psi_{\mathrm{c}}\left(\mathrm{concat}\left(\gamma\left(\mathbf{x}_{\mathrm{c}}\right),\,\mathbf{O}_{\mathrm{c}}^{i}\right)\right)\longmapsto\left(\mathbf{c},\,\sigma\right)\,.
\end{equation}

\subsection{Training}

\noindent{\bf Rendering.} To render the pixel color, we shoot two rays ($\mathbf{r}_{\mathrm{s}}, \mathbf{r}_{\mathrm{ho}}$) and sample intervals (i.e., ray points) to query the scene model (using $\mathbf{r}_{\mathrm{s}}$) and dynamic human-object model (with $\mathbf{r}_{\mathrm{ho}}$) respectively in their coordinate systems. After aligning the coordinates (see supplementary materials), we transform the sampled 3D points and their queried properties from the dynamic human-object space to the scene space, and re-order all sampled properties based on their distance from the camera center, as shown in Fig.~\ref{fig:pipeline}. Therefore, the color of a pixel can be calculated through volume rendering, \emph{i.e.}, by integrating the re-ordered properties along the ray~\cite{mildenhall2021nerf}:
\begin{eqnarray}\label{eq:render}
\hat{\mathbf{C}}\left(\mathbf{r}\right) =  \sum_{i=1}^{N}T_{i}\left(1-e^{-\sigma_{i}\delta_{i}}\right)\mathbf{c}_{i}\,,T_{i}  = e^{-\sum_{j=1}^{i-1}\sigma_{j}\delta_{j}}\,.
\end{eqnarray}

Directly fusing $\mathbf{r}_{\mathrm{s}}$ and $\mathbf{r}_{\mathrm{ho}}$ may yield minor artifacts on overlapped ground regions due to intersections of human-object points and scene points on short-length ground rays. To solve this issue, we estimate the foreground mask during training to separate the background rays from foreground ones. This allows us to focus on combining $\mathbf{r}_{\mathrm{s}}$ and $\mathbf{r}_{\mathrm{ho}}$ at the overlapped foreground regions, thereby reducing artifacts.

\noindent{\bf Training Objectives.}
The training of \method{} consists of three stages. In the first stage, we mask out the dynamic human-object regions and train the state-conditional scene model. Then, we train the state-conditional dynamic human-object model on the dynamic human-object regions. In the third stage, we combine these two models and further finetune the complete HOSNeRF for all image pixels.

Given the single video with calibrated poses, the first stage of \method{} is optimized by minimizing the photometric MSE loss and the regularization losses proposed by Mip-NeRF 360~\cite{barron2022mip} to avoid background collapse. In the second and third stages of \method{}, we utilize the photometric MSE loss, patched-based perceptual LPIPS~\cite{zhang2018unreasonable} loss, forward-backward deformation cycle consistency, and indirect optical flow supervisions.

The deformation cycle consistency is enforced by minimizing the distance between estimated deformed points of Eq. (\ref{eq:cycle_func}) and sampled deformed points $\mathbf{x}_{\mathrm{d}}$,
\begin{equation}
\mathcal{L}_{\mathrm{Cycle}}=\frac{1}{2N}\sum_{i=1}^{N}\left\Vert \mathbf{x}_{\mathrm{d}}^{i}-\mathbf{\hat{x}}_{\mathrm{d}}^{i}\right\Vert _{2}^{2}\,.\label{eq:cycle}
\end{equation}

We also incorporate 2D optical flow as indirect supervision by minimizing the error between the induced flow and the estimated flow (details in the supplementary material). Therefore, the overall training objective of \method{} is:
\begin{eqnarray}
\mathcal{L} & = & \omega_{\mathrm{MSE}}\cdot\mathcal{L}_{\mathrm{MSE}}+\omega_{\mathrm{LPIPS}}\mathcal{L}_{\mathrm{LPIPS}}\nonumber \\
 & + & \omega_{\mathrm{Cycle}}\cdot\mathcal{L}_{\mathrm{Cycle}}+\omega_{\mathrm{Flow}}\cdot\mathcal{L}_{\mathrm{Flow}}\,,
\end{eqnarray}
\noindent where $\omega_{\mathrm{MSE}},\,\omega_{\mathrm{LPIPS}},\,\omega_{\mathrm{Cycle}},\,\omega_{\mathrm{Flow}}$ are loss weights.

\definecolor{myyellow}{rgb}{1,1, 0.6}
\definecolor{myorange}{rgb}{1, 0.8, 0.6}
\definecolor{myred}{rgb}{1, 0.6, 0.6}

\section{Experiments}

\subsection{Dataset}

\begin{table}

\begin{centering}
\begin{tabular}{ccc}
\hline 
Scene & No. of objects & No. of states\tabularnewline
\hline 
\textsc{\small Backpack } & 1 & 3\tabularnewline
\textsc{\small Tennis } & 2 & 3\tabularnewline
\textsc{\small Suitcase } & 1 & 4\tabularnewline
\textsc{\small Playground } & 3 & 5\tabularnewline
\textsc{\small Dance } & 3 & 7\tabularnewline
\textsc{\small Lounge } & 3 & 5\tabularnewline
\hline 
\end{tabular}
\par\end{centering}
\vspace{-2mm}
\caption{Details of our collected dataset.
\label{tab:dataset}
}
\vspace{-5mm}
\end{table}

To benchmark the reconstruction of monocular videos with dynamic human-object-scenes, we collect a new dataset with 6 scenes\footnote{Our data collection process has been approved by the institutional review board and participants have given consents for public data release.}. The collected dataset features various types of human-object-scene interactions in indoor and outdoor scenarios, with up to $3$ interacted objects and $7$ states for a single video, as reported in Tab.~\ref{tab:dataset}. The duration of collected videos varies from $60\mathrm{s}$ to $120\mathrm{s}$, and we extract $\left[300,\,400\right]$ frames for each video, where 16 frames at equal intervals are selected as novel views at novel timesteps as the test set and the remaining frames are the train set. 

We extensively evaluate the \method{} on our collected dataset and NeuMan dataset~\cite{jiang2022neuman} that consists of 6 short human walking sequences. Since NeuMan dataset does not involve human-object interactions, \method{} can be flexibly customized to HSNeRF by removing object bones and setting the object state to $1$ when evaluated on NeuMan dataset.

\subsection{Implementation Details}

Human interacts with various objects using hands in our dataset, we therefore introduce $2$ new object bones to the left and right hand joints of the 3D human skeleton for all scenes. As a result, there are in total $K=26$ bones in our human-object skeleton. In order to obtain the human-object pose prior, we first utilize a pretrained ROMP~\cite{sun2021monocular} model to estimate the 3D human pose, and use Mask-RCNN~\cite{he2017mask} to estimate human and objects masks. However, since it is non-trivial to estimate the object pose for in-the-wild videos, we initialize the left and right object bones as the standard extended bones with no relative rotations with respect to human hands, and initialize their length by referencing the length of the object relative to the arm. During training, we further refine the human-object poses $\left(\mathcal{J},\,\mathcal{R}\right)$ with a MLP-based pose correction module $\left(\Psi_{\mathcal{J}},\,\Psi_{\mathcal{R}}\right)$~\cite{weng2022humannerf}. The overall three-stage training of \method{} takes about $5$ days on $4$ Tesla V100 GPUs. Please see the supplementary material for the comparison of training time with other approaches and discussions about estimating object masks.

\subsection{Experimental Results}
\noindent\textbf{Baselines and Evaluation Metrics.} To demonstrate the performance of \method{}, we make comparisons with various types of SOTA approaches, including (1) NeuMan~\cite{jiang2022neuman}, a human pose-based method; (2) HyperNeRF~\cite{park2021hypernerf} and (3) Nerfies~\cite{park2021nerfies}, dynamic radiance fields for general scenes; (4) D$^2$NeRF~\cite{wu2022d}, a static-dynamic decomposition method; (5) K-Planes~\cite{fridovich2023k}, an explicit radiance field method. For quantitative comparison, peak signal-to-noise ratio (PSNR), structural similarity index (SSIM)~\cite{wang2004image}, and Learned Perceptual Image Patch Similarity (LPIPS)~\cite{zhang2018unreasonable} are employed as evaluation metrics.

\begin{figure*}
\begin{centering}
\includegraphics[width=0.99\linewidth]{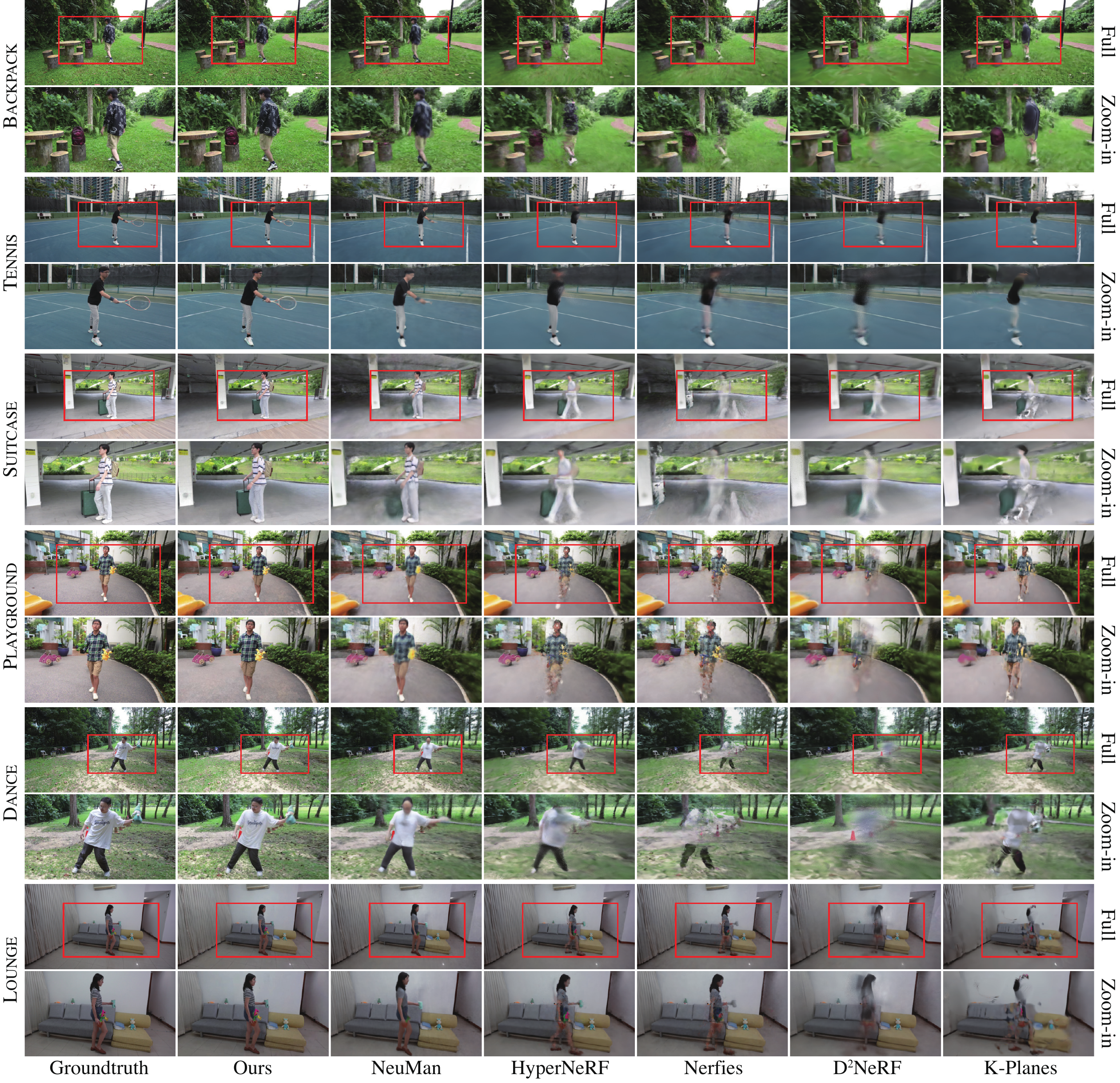}
\par\end{centering}
\vspace{-2mm}
\caption{\label{fig:vis_result}Qualitative comparisons of \method{} and SOTA approaches on \method{} dataset.}
\vspace{-5mm}
\end{figure*}

\newcommand{\tablefirst}[0]{\cellcolor{myred}}
\newcommand{\tablesecond}[0]{\cellcolor{myorange}}
\newcommand{\tablethird}[0]{\cellcolor{myyellow}}

\begin{table*}[h]
\centering

\resizebox{\linewidth}{!}{
\centering
\setlength{\tabcolsep}{1.8pt}

\begin{tabular}{l||ccc||ccc||ccc||ccc||ccc||ccc}

\toprule
& \multicolumn{ 3 }{c||}{
  \makecell{
  \textsc{\small Backpack }
  }
}
& \multicolumn{ 3 }{c||}{
  \makecell{
  \textsc{\small Tennis }
  }
}
& \multicolumn{ 3 }{c||}{
  \makecell{
  \textsc{\small Suitcase }
  }
}
& \multicolumn{ 3 }{c||}{
  \makecell{
  \textsc{\small Playground }
  }
}
& \multicolumn{ 3 }{c||}{
  \makecell{
  \textsc{\small Dance }
  }
}
& \multicolumn{ 3 }{c}{
  \makecell{
  \textsc{\small Lounge }
  }
}
\\

& \multicolumn{1}{c}{ \footnotesize PSNR$\uparrow$ }
& \multicolumn{1}{c}{ \footnotesize SSIM$\uparrow$ }
& \multicolumn{1}{c||}{ \footnotesize LPIPS$\downarrow$ }
& \multicolumn{1}{c}{ \footnotesize PSNR$\uparrow$ }
& \multicolumn{1}{c}{ \footnotesize SSIM$\uparrow$ }
& \multicolumn{1}{c||}{ \footnotesize LPIPS$\downarrow$ }
& \multicolumn{1}{c}{ \footnotesize PSNR$\uparrow$ }
& \multicolumn{1}{c}{ \footnotesize SSIM$\uparrow$ }
& \multicolumn{1}{c||}{ \footnotesize LPIPS$\downarrow$ }
& \multicolumn{1}{c}{ \footnotesize PSNR$\uparrow$ }
& \multicolumn{1}{c}{ \footnotesize SSIM$\uparrow$ }
& \multicolumn{1}{c||}{ \footnotesize LPIPS$\downarrow$ }
& \multicolumn{1}{c}{ \footnotesize PSNR$\uparrow$ }
& \multicolumn{1}{c}{ \footnotesize SSIM$\uparrow$ }
& \multicolumn{1}{c||}{ \footnotesize LPIPS$\downarrow$ }
& \multicolumn{1}{c}{ \footnotesize PSNR$\uparrow$ }
& \multicolumn{1}{c}{ \footnotesize SSIM$\uparrow$ }
& \multicolumn{1}{c}{ \footnotesize LPIPS$\downarrow$ }
\\
\hline

  K-Planes~\cite{fridovich2023k}
  &$19.05$
  &$0.345$
  &$0.557$
  
  &$19.31$
  &$0.690$
  &$0.536$
  
  &$18.64$
  &$0.575$
  &$0.602$

  &$17.92$
  &$0.446$
  &$0.635$
  
  &$18.17$
  &$0.328$
  &$0.623$
  
  &$24.21$
  &$0.824$
  &$0.453$

  \\ 
  D$^2$NeRF~\cite{wu2022d}
  &$20.52$
  &$0.384$
  &$0.608$
  
  &$23.97$
  &$0.737$
  &$0.540$
  
  &$20.99$
  &$0.592$
  &$0.645$

  &$21.23$
  &$0.510$
  &$0.616$
  
  &$19.92$
  &$0.407$
  &$0.647$
  
  &$27.13$
  &$0.858$
  &$0.509$
  
  \\ 
  Nerfies~\cite{park2021nerfies}
  &$19.56$
  &$0.515$
  &$0.559$
  
  &$22.12$
  &$0.818$
  &$0.443$
  
  &$19.01$
  &$0.643$
  &$0.555$

  &$21.14$
  &$0.726$
  &$0.533$
  
  &$19.37$
  &$0.674$
  &$0.524$
  
  &$25.90$
  &$0.914$
  &$0.342$
  
  \\ 
  HyperNeRF~\cite{park2021hypernerf}
  &$19.62$
  &$0.359$
  &$0.587$
  
  &$21.26$
  &$0.742$
  &$0.510$
  
  &$19.41$
  &$0.604$
  &$0.607$

  &$21.67$
  &$0.558$
  &$0.578$
  
  &$19.30$
  &$0.412$
  &$0.601$
  
  &$27.25$
  &$0.925$
  &$0.332$
  
  \\  
  NeuMan~\cite{jiang2022neuman}
  &$21.21$
  &$0.479$
  &$0.478$
  
  &$23.17$
  &$0.768$
  &$0.442$
  
  &$20.84$
  &$0.611$
  &$0.551$

  &$21.46$
  &$0.546$
  &$0.551$
  
  &$21.19$
  &$0.529$
  &$0.490$
  
  &\tablefirst$28.40$
  &$0.917$
  &$0.341$

  \\ 
  \hline  
  Ours (base)
  &$21.51$
  &$0.764$
  &$0.271$
  
  &$24.02$
  &$0.910$
  &$0.326$
  
  &$21.10$
  &$0.829$
  &$0.395$

  &$22.20$
  &$0.787$
  &$0.348$
  
  &$21.84$
  &$0.785$
  &$0.266$
  
  &$26.54$
  &$0.962$
  &$0.243$

  \\    
  Ours w/ state
  &$22.38$
  &$0.786$
  &$0.252$
  
  &$23.98$
  &$0.910$
  &$0.323$
  
  &$21.43$
  &$0.834$
  &$0.390$

  &$22.52$
  &$0.796$
  &$0.341$
  
  &$22.43$
  &$0.796$
  &$0.258$
  
  &$27.51$
  &$0.965$
  &$0.245$

  \\    
  Ours w/ object
  &$21.56$
  &$0.767$
  &$0.269$
  
  &$23.98$
  &$0.909$
  &$0.327$
  
  &$21.08$
  &$0.830$
  &$0.396$

  &$22.14$
  &$0.787$
  &$0.347$
  
  &$21.88$
  &$0.785$
  &$0.269$
  
  &$26.80$
  &$0.962$
  &$0.246$

  \\    
  Ours w/ state, object
  &$22.33$
  &$0.785$
  &$0.253$
  
  &$24.08$
  &$0.909$
  &\tablesecond$0.320$
  
  &$21.51$
  &$0.833$
  &$0.386$

  &$22.55$
  &$0.796$
  &$0.338$
  
  &$22.38$
  &$0.796$
  &$0.259$
  
  &$27.52$
  &$0.965$
  &$0.245$

  \\    
  Ours w/ state, object, mask
  &\tablesecond$22.48$
  &\tablesecond$0.790$
  &\tablesecond$0.245$
  
  &\tablefirst$24.19$
  &\tablesecond$0.911$
  &$0.321$
  
  &\tablesecond$21.60$
  &\tablesecond$0.834$
  &\tablesecond$0.382$

  &\tablefirst$22.77$
  &\tablesecond$0.799$
  &\tablefirst$0.335$
  
  &\tablesecond$22.48$
  &\tablesecond$0.802$
  &\tablesecond$0.251$
  
  &$27.73$
  &\tablesecond$0.968$
  &\tablesecond$0.227$

  \\    
  Ours (full)
  &\tablefirst$22.56$
  &\tablefirst$0.792$
  &\tablefirst$0.243$
  
  &\tablesecond$24.15$
  &\tablefirst$0.911$
  &\tablefirst$0.320$
  
  &\tablefirst$21.74$
  &\tablefirst$0.836$
  &\tablefirst$0.382$

  &\tablesecond$22.67$
  &\tablefirst$0.801$
  &\tablesecond$0.336$
  
  &\tablefirst$22.63$
  &\tablefirst$0.804$
  &\tablefirst$0.248$
  
  &\tablesecond$27.74$
  &\tablefirst$0.968$
  &\tablefirst$0.227$
  
  \\ \bottomrule

\end{tabular}
}

\vspace{-2mm}
\caption{
Per-scene quantitative evaluation on the HOSNeRF dataset against baselines and ablations of our method. We color code each cell as \colorbox{myred}{\textbf{best}} and \colorbox{myorange}{\textbf{second best}}. 
\label{tab:hosnerf_comparison}
}
\vspace{-1mm}
\end{table*}

\noindent\textbf{Evaluation on \method{} Dataset.} We report the quantitative comparisons (PSNR, SSIM, and LPIPS with VGG~\cite{simonyan2014very} backbone) of $6$ scenes for all approaches in Tab.~\ref{tab:hosnerf_comparison}. For all SOTA approaches, we utilize their highest configurations for a fair comparison. As shown in Tab.~\ref{tab:hosnerf_comparison}, our~\method{} achieves the best performance in terms of all metrics, except the PSNR for the Lounge scene (a metric known to favor smooth/blurry results~\cite{zhang2018unreasonable}). The improvement of~\method{} is particularly significant in terms of the LPIPS, with an average of nearly $40\%$ gain over SOTA approaches. Fig.~\ref{fig:vis_result} visualizes the qualitative comparison of \method{} over SOTA approaches on novel views at novel timesteps, where \method{} achieves substantially better visual quality than other approaches for all scenes. \method{} is able to produce high-fidelity details close to ground truths for all scene contents, i.e., dynamic human bodies, objects, and backgrounds. In contrast, existing approaches tend to synthesize much blurrier images with missing components. Please see the supplementary video for more results on 360{\textdegree} free-viewpoint bullet-time videos rendered by \method{}.

\begin{table*}
\centering

\resizebox{\linewidth}{!}{
\centering
\setlength{\tabcolsep}{1.8pt}

\begin{tabular}{l||ccc||ccc||ccc||ccc||ccc||ccc}

\toprule
& \multicolumn{ 3 }{c||}{
  \makecell{
  \textsc{\small Seattle }
  }
}
& \multicolumn{ 3 }{c||}{
  \makecell{
  \textsc{\small Parking }
  }
}
& \multicolumn{ 3 }{c||}{
  \makecell{
  \textsc{\small Bike }
  }
}
& \multicolumn{ 3 }{c||}{
  \makecell{
  \textsc{\small Jogging }
  }
}
& \multicolumn{ 3 }{c||}{
  \makecell{
  \textsc{\small Lab }
  }
}
& \multicolumn{ 3 }{c}{
  \makecell{
  \textsc{\small Citron }
  }
}
\\

& \multicolumn{1}{c}{ \footnotesize PSNR$\uparrow$ }
& \multicolumn{1}{c}{ \footnotesize SSIM$\uparrow$ }
& \multicolumn{1}{c||}{ \footnotesize LPIPS$\downarrow$ }
& \multicolumn{1}{c}{ \footnotesize PSNR$\uparrow$ }
& \multicolumn{1}{c}{ \footnotesize SSIM$\uparrow$ }
& \multicolumn{1}{c||}{ \footnotesize LPIPS$\downarrow$ }
& \multicolumn{1}{c}{ \footnotesize PSNR$\uparrow$ }
& \multicolumn{1}{c}{ \footnotesize SSIM$\uparrow$ }
& \multicolumn{1}{c||}{ \footnotesize LPIPS$\downarrow$ }
& \multicolumn{1}{c}{ \footnotesize PSNR$\uparrow$ }
& \multicolumn{1}{c}{ \footnotesize SSIM$\uparrow$ }
& \multicolumn{1}{c||}{ \footnotesize LPIPS$\downarrow$ }
& \multicolumn{1}{c}{ \footnotesize PSNR$\uparrow$ }
& \multicolumn{1}{c}{ \footnotesize SSIM$\uparrow$ }
& \multicolumn{1}{c||}{ \footnotesize LPIPS$\downarrow$ }
& \multicolumn{1}{c}{ \footnotesize PSNR$\uparrow$ }
& \multicolumn{1}{c}{ \footnotesize SSIM$\uparrow$ }
& \multicolumn{1}{c}{ \footnotesize LPIPS$\downarrow$ }
\\
\hline

  NSFF~\cite{li2021neural}
  &$21.84$
  &$0.69$
  &$0.37$
  
  &$21.98$
  &$0.69$
  &$0.46$
  
  &$21.16$
  &$0.71$
  &$0.36$

  &$20.63$
  &$0.53$
  &$0.49$
  
  &$20.52$
  &$0.75$
  &$0.39$
  
  &$12.33$
  &$0.49$
  &$0.65$
  
  \\ 
  HyperNeRF~\cite{park2021hypernerf}
  &$16.43$
  &$0.43$
  &$0.40$
  
  &$16.04$
  &$0.38$
  &$0.62$
  
  &$17.64$
  &$0.42$
  &$0.43$

  &$18.52$
  &$0.39$
  &$0.52$
  
  &$16.75$
  &$0.51$
  &$0.23$
  
  &$16.81$
  &$0.41$
  &$0.56$
  
  \\  
  NeuMan~\cite{jiang2022neuman}
  &$23.98$
  &$0.77$
  &$0.26$
  
  &$25.43$
  &$0.79$
  &$0.31$
  
  &$25.52$
  &$0.82$
  &$0.23$

  &$22.68$
  &$0.67$
  &$0.32$
  
  &$24.93$
  &$0.85$
  &$0.21$
  
  &\tablefirst$24.71$
  &$0.80$
  &$0.26$

  \\ 
  \hline  
  Ours
  &\tablefirst$26.68$
  &\tablefirst$0.91$
  &\tablefirst$0.14$
  
  &\tablefirst\tablefirst$27.20$
  &\tablefirst$0.93$
  &\tablefirst$0.12$
  
  &\tablefirst$26.09$
  &\tablefirst$0.93$
  &\tablefirst$0.10$

  &\tablefirst$25.04$
  &\tablefirst$0.89$
  &\tablefirst$0.16$
  
  &\tablefirst$24.93$
  &\tablefirst$0.94$
  &\tablefirst$0.10$
  
  &$24.44$
  &\tablefirst$0.90$
  &\tablefirst$0.18$
  
  \\ \bottomrule

\end{tabular}
}

\vspace{-2mm}
\caption{
Per-scene quantitative evaluation on the NeuMan dataset against baselines. We color code each cell as \colorbox{myred}{\textbf{best}}. 
\label{tab:neuman_comparison}
}
\vspace{-4mm}
\end{table*}

\noindent\textbf{Evaluation on NeuMan Dataset~\cite{jiang2022neuman}.} We report quantitative metrics (PSNR, SSIM, and LPIPS with AlexNet~\cite{krizhevsky2017imagenet} backbone) over $6$ scenes of the NeuMan dataset in Tab.~\ref{tab:neuman_comparison}, where the metrics for NSFF~\cite{li2021neural}, HyperNeRF~\cite{park2021hypernerf}, and NeuMan~\cite{jiang2022neuman} are borrowed from NeuMan~\cite{jiang2022neuman}. As shown in Tab.~\ref{tab:neuman_comparison}, our~\method{} achieves the best performance in terms of all metrics except the PSNR for the Citron scene, and significantly improves SOTA approaches by a large margin of $50\%$ in terms of LPIPS. This further demonstrates the effectiveness and flexibility of \method{} to model various types of dynamic human-(objects)-scenes. Please see the supplementary material for qualitative comparison.

\subsection{Ablation Study}

\begin{figure}
\begin{centering}
\includegraphics[width=1\linewidth]{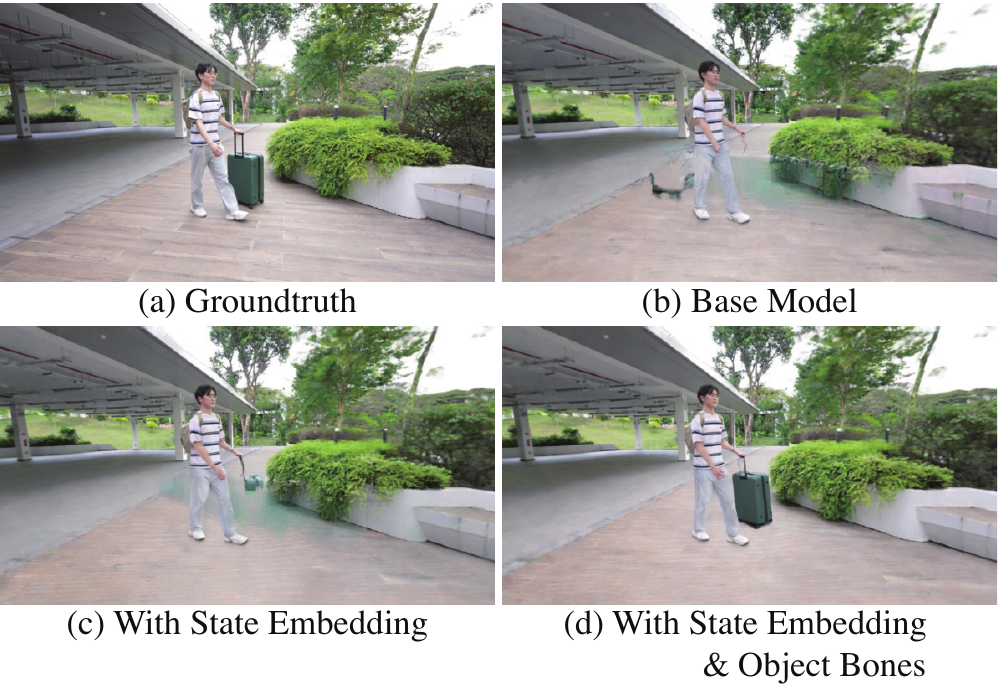}
\par\end{centering}
\vspace{-2mm}
\caption{\label{fig:Ablation_suitcase}Ablation of \method{} for large objects.}
\vspace{-1mm}
\end{figure}

\begin{figure}
\begin{centering}
\includegraphics[width=1\linewidth]{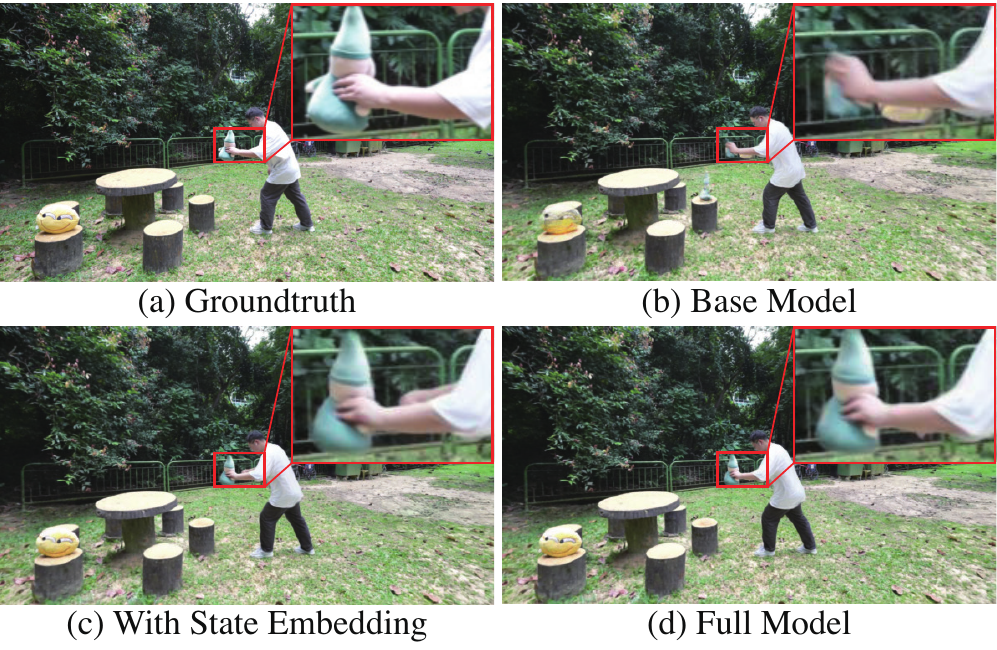}
\par\end{centering}
\vspace{-2mm}
\caption{\label{fig:Ablation_dance}Ablation of \method{} for small objects.}
\vspace{-6mm}
\end{figure}

We conduct ablation studies on our collected dataset to evaluate the effectiveness of each proposed component in \method{}. We progressively ablate each component from state embeddings, object bones, foreground masks, optical flow, and cycle consistency supervision. As shown in Tab.~\ref{tab:hosnerf_comparison}, the performance of \method{} progressively drops with the disabling of each component, with the full model nearly achieving the best performances, which demonstrates the effectiveness of our designs.

We visualize representative scenes with large (Fig.~\ref{fig:Ablation_suitcase}) and small (Fig.~\ref{fig:Ablation_dance}) objects to further evaluate our proposed designs. As shown in Fig.~\ref{fig:Ablation_suitcase}(b) and \ref{fig:Ablation_dance}(b), the base models are unable to represent the interacted objects for the human and the scene. In contrast, the proposed state embedding is capable of representing the state transitions of small objects (Fig.~\ref{fig:Ablation_dance}(c)), but it fails to represent the deformations of large objects (Fig.~\ref{fig:Ablation_suitcase}(c)). Together with our proposed object bones, the deformation of large objects can be effectively estimated through the human-object skeleton hierarchy, enabling large objects modeling in Fig.~\ref{fig:Ablation_suitcase}(d). In addition, the full model can further improve the synthesis quality of fine details such as correcting the wrong right arm by comparing Fig.~\ref{fig:Ablation_dance}(c) and (d). These comparisons further demonstrate the importance and effectiveness of the proposed designs.

\section{Conclusion}

We introduced a novel framework of \method{}, the first work to achieve 360{\textdegree} free-viewpoint high-fidelity novel view synthesis for dynamic scenes with human-environment interactions from a single video. To tackle the challenges of such a task, we first introduced the new object bones to the human skeleton hierarchy to effectively estimate the fast human-object motion in our dynamic human-object model. Then, we proposed the state-conditional human-object and scene representations for handling significant object state changes. With our designed training strategies, \method{} produced significant improvements over SOTA approaches, and enabled 360{\textdegree} free-viewpoint rendering of any frame from a single video and supported rendering views with novel human and object poses.

\noindent\textbf{Limitations and Future Work.} Our \method{} focuses on dynamic human-objects modeling and lacks the capability to represent dynamic backgrounds. It remains challenging but is worthwhile researching models for reconstructing high-fidelity dynamic unbounded backgrounds.

{\small
\bibliographystyle{ieee_fullname}
\bibliography{egbib}
}

\appendix
\newpage

\section*{Appendix}
The supplementary material is structured as follows:

\begin{itemize}
\item Sec.~\ref{sec:1} provides further implementation details of the proposed \method{}.
\item Sec.~\ref{sec:2} presents additional details on the network designs of our \method{}.
\item Sec.~\ref{sec:3} summarizes additional comparisons of our \method{} against state-of-the-art (SOTA) approaches.
\end{itemize}

Furthermore, we also provide a \textbf{supplementary video} showcasing per-scene 360{\textdegree} free-viewpoint renderings from our \method{} on all six scenes of our \method{} dataset.

\section{Implementation Details}
\label{sec:1}

We conducted all our experiments on 4 Tesla V100 GPUs, using the PyTorch~\cite{paszke2019pytorch} deep learning framework.

\noindent\textbf{Optical Flow Supervision.} 
We first map the deformed points $\mathbf{x}_{\mathrm{d}}$ from the deformed space at timestep $t$ to canonical points $\mathbf{x}_{\mathrm{c}}$ in the canonical space. Then we compute their corresponding deformed points at timestep $t-1$, denoted as $\mathbf{\hat{x}}_{\mathrm{d}_{t-1}}$, through forward deformation:
\begin{equation}
\mathbf{\hat{x}}_{\mathrm{d}_{t-1}}=\Psi_{\mathrm{c}\rightarrow\mathrm{d}_{t-1}}^{\mathrm{coarse}}\left(\mathbf{x}_{\mathrm{c}},\,\mathcal{J},\,\mathcal{R}\right)+\Delta\mathbf{x}_{\mathrm{c}\rightarrow\mathrm{d}_{t-1}}.\,
\end{equation}
We project $\hat{\mathcal{X}}_{\mathrm{d}_{t-1}}=\left\{\mathbf{\hat{x}}_{\mathrm{d}_{t-1}}^{i}\right\}$ onto the reference camera at timestep $t-1$ to obtain their corresponding pixel locations $\hat{\mathcal{P}}_{\mathrm{d}_{t-1}}=\{\hat{\mathbf{P}}_{\mathrm{d}_{t-1}}^{i}\}$. We then compute the optical flow induced by these points with respect to the pixel locations ${\mathcal{P}}_{\mathrm{d}_{t}}=\{{\mathbf{P}}_{\mathrm{d}_{t}}^{i}\}$ from which the rays of $\mathcal{X}_{\mathrm{d}}=\left\{\mathbf{x}_{\mathrm{d}}^{i}\right\}$ are cast. Finally, we minimize the error between the induced flow and the estimated flow:
\begin{equation}
\mathcal{L}_{\mathrm{Flow}}=\frac{1}{\left|\mathcal{R}\right|}\sum_{\mathbf{r}\in\mathcal{R}}\sum_{i=1}^{N}w^{\mathbf{r},i}\left|\left|\left(\hat{\mathbf{P}}_{\mathrm{d}_{t-1}}^{\mathbf{r},i}-\mathbf{P}_{\mathrm{d}_{t}}^{\mathbf{r},i}\right)-\mathbf{f}_{\mathbf{P}_{\mathrm{d}_{t}}^{\mathbf{r},i}}\right|\right|\,,
\end{equation}
\noindent where $w^{\mathbf{r},i}=T_{i}\left(1-\mathrm{exp}\left(-\sigma_{i}\delta_{i}\right)\right)$
is the ray termination weights from the volume rendering equation,
and $\mathbf{f}_{\mathbf{P}_{\mathrm{d}_{t}}^{\mathbf{r},i}}$ is the estimated 2D backward optical
flow using RAFT~\cite{teed2020raft} at  $\mathbf{P}^{t}_{\mathbf{r},i}$.

\noindent\textbf{Coordinate System Alignment.} 
To integrate the state-conditional scene model and dynamic human-object model, we initially synchronize their coordinate systems during preprocessing, as they are originally processed and defined in separate coordinate systems. To achieve this, we utilize the SMPL~\cite{loper2015smpl} parameters acquired from the pre-trained human pose estimation model ROMP~\cite{sun2021monocular} and adopt the scene-SMPL alignment approach from NeuMan~\cite{jiang2022neuman}. This technique requires that the human subject always stands on the ground. Subsequently, we align the two coordinate systems through the Perspective-n-Point (PnP)~\cite{lepetit2009epnp} method and resolve any scale ambiguities by restricting the feet meshes of the SMPL model to touch the ground plane~\cite{jiang2022neuman}. In this context, the near and far parameters for the scene model are set to $0.1$ and $10^6$, respectively, while those for the dynamic human-object model are determined by the coarse bounding box calculated from the human-object poses.

\noindent\textbf{Human and Object Masks.} To estimate human and object masks, we utilize the pre-trained Mask-RCNN~\cite{he2017mask} model. Consequently, the majority of object classes in our dataset come from the COCO~\cite{lin2014microsoft} dataset. During preprocessing, we successfully segment all humans and most objects in our dataset. However, for objects that are not detected due to occlusions or out-of-domain classes, we manually segment them. To ensure complete separation of the foreground from the background, we then dilate the human and object masks by $5\%$. The proposed three-stage training pipeline of our \method{} method is beneficial, especially the third stage, which involves fine-tuning for foreground-background merging. This enables training with {\it coarse} human and object masks. In contrast, HumanNeRF~\cite{weng2022humannerf} depends on manual intervention to correct coarse segmentation errors.

\noindent\textbf{Optimization Parameters.} We optimize our \method{} using Adam optimizer~\cite{kingma2014adam}. We set the base learning rates for our training process as follows: $0.002$ for the first stage to train the background, $0.0006$ for the second stage to train the dynamic human-object model, and $0.00006$ for the third stage to fine-tune the complete \method{} model. For most of the scenes, we balance the loss terms using 
 the following weighting factors: $\omega_{\mathrm{MSE}}=0.2,\,\omega_{\mathrm{LPIPS}}=1.0,\,\omega_{\mathrm{Cycle}}=0.01,\,\omega_{\mathrm{Flow}}=0.01$. The three stages are trained for $500\mathrm{k}$, $400\mathrm{k}$, and $200\mathrm{k}$ iterations, respectively.

\begin{figure}
\begin{centering}
\includegraphics[width=1\linewidth]{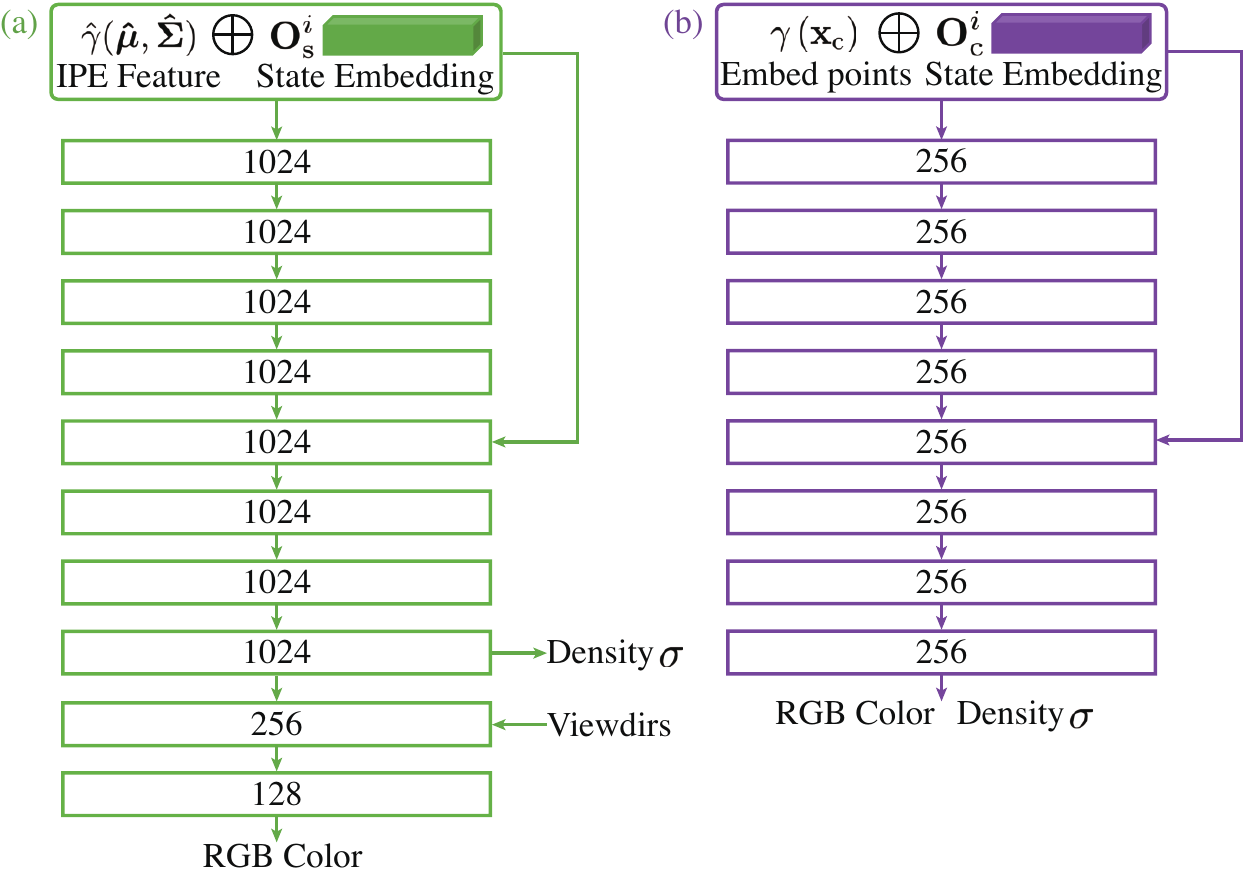}
\par\end{centering}
\vspace{-1mm}
\caption{\label{fig:state_networks}State-conditional network designs for the scene base model (a) and the canonical space model (b).}
\vspace{-3mm}
\end{figure}

\section{Network Details}
\label{sec:2}
\noindent\textbf{Object State Embeddings.} To address the issue of humans interacting with different objects at different times, we introduce two new learnable object state embeddings that serve as conditions for learning our human-object representation and scene representation, respectively. In a dynamic scene with $N$ object states, we define $N$ learnable state embeddings $\mathcal{O}_{\mathrm{s}}=\left\{ \mathbf{O}_{\mathrm{s}}^{i}\right\} \,\left(i=1,2,\cdots,N\right)$ to represent object states in the scene model, and $N$ learnable state embeddings $\mathcal{O}_{\mathrm{c}}=\left\{ \mathbf{O}_{\mathrm{c}}^{i}\right\} \,\left(i=1,2,\cdots,N\right)$ to represent object states in the canonical space.  The feature dimension of $\mathcal{O}_{\mathrm{s}}$ and $\mathcal{O}_{\mathrm{c}}$ are both set to $64$ in our model. To obtain the number of object states, we manually label the transition timesteps for each video when the human picks up or puts down objects. Alternatively, we could use pretrained affordance detection methods to detect these transition timesteps. In our newly collected dataset, we provide the ground-truth transition timesteps for all the scenes.

\noindent\textbf{State-Conditional Scene Network.} As shown in Fig.~\ref{fig:state_networks}(a), we employ a 10-layer multilayer perceptron (MLP) as our state-conditional scene base network, following the approach outlined in Mip-NeRF 360~\cite{barron2022mip}. Specifically, at state $i$, we utilize a concatenation of the IPE features $\hat{\gamma}(\boldsymbol{\hat{\mu}}, \boldsymbol{\hat{\Sigma}})$ of ray intervals with the scene state embedding $\mathbf{O}_{\mathrm{s}}^{i}$ as input to the scene MLP. To achieve this, we employ a skip connection that concatenates the input to the fifth layer. For the activation functions, we use ReLU after each fully connected layer, except for predicting density, for which we use Softplus, and for predicting color, for which we use Sigmoid.

\noindent\textbf{State-Conditional Canonical Space Network.} As illustrated in Fig.~\ref{fig:state_networks}(b), we follow NeRF~\cite{mildenhall2021nerf} to use an 8-layer MLP as our state-conditional canonical space model. At object state $i$, we concatenate the positionally encoded canonical points $\gamma\left(\mathbf{x}_{\mathrm{c}}\right)$ with the human-object state embedding $\mathbf{O}_{\mathrm{c}}^{i}$ and pass them to the canonical space MLP. In this canonical MLP, we adopt a skip connection that concatenates the input to the fifth layer. We use the ReLU activation after each fully connected layer, with the exception of the prediction of color, for which we employ the Sigmoid activation function.

\section{Additional Results}
\label{sec:3}

\begin{figure*}
\begin{centering}
\includegraphics[width=1\linewidth]{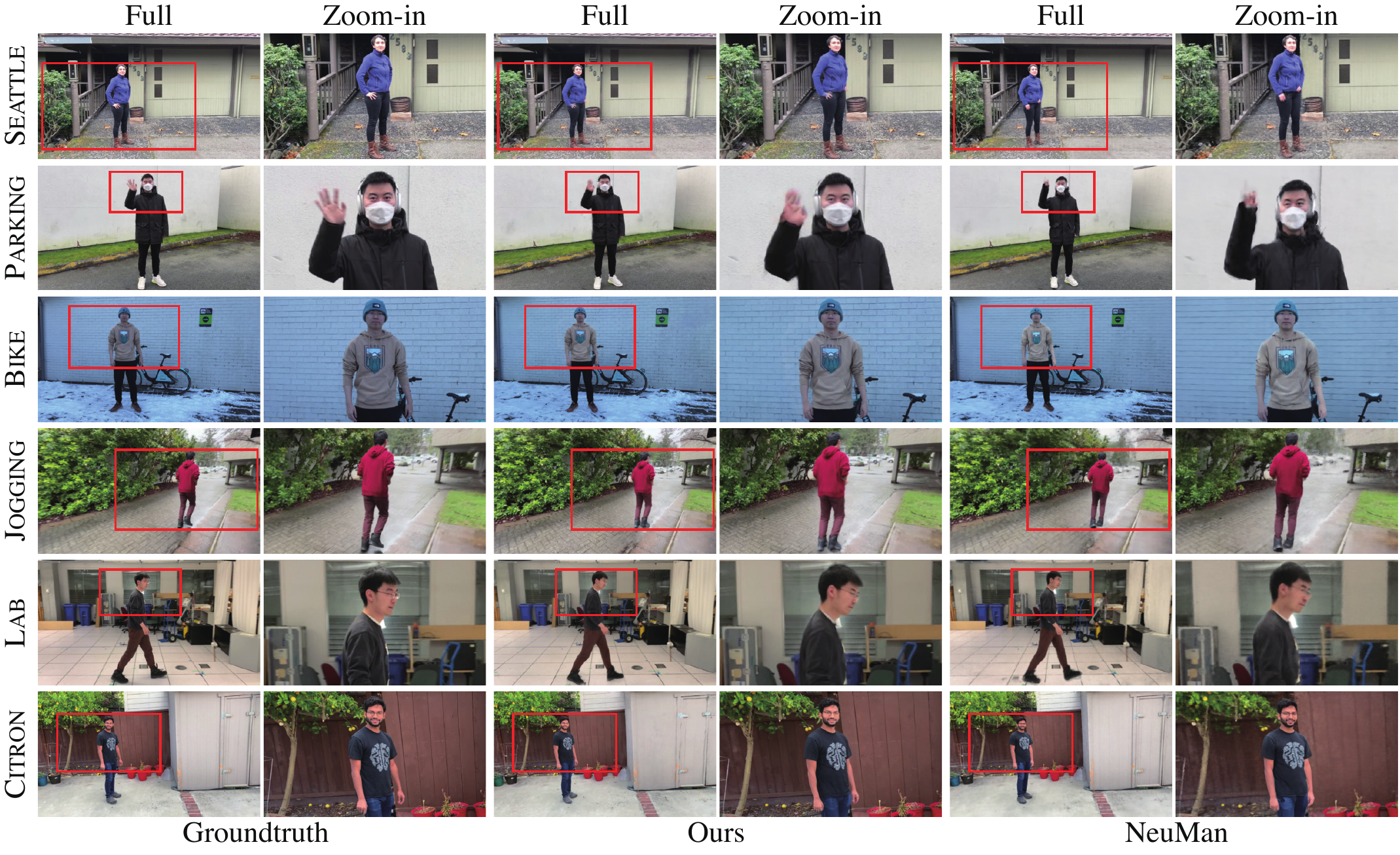}
\par\end{centering}
\caption{\label{fig:vis_result_neuman}Qualitative comparisons of \method{} and NeuMan~\cite{jiang2022neuman} on the NeuMan dataset.}
\end{figure*}

\noindent\textbf{Qualitative Comparisons on NeuMan Dataset~\cite{jiang2022neuman}.} NeuMan dataset~\cite{jiang2022neuman} consists of 6 short human walking sequences varying from $10\mathrm{s}$ to $20\mathrm{s}$. Since the NeuMan dataset does not involve human-object interactions, \method{} can be flexibly customized to HSNeRF by removing object bones and setting the object state to $1$ when evaluated on the NeuMan dataset. Fig.~\ref{fig:vis_result_neuman} visualizes the qualitative comparisons between \method{} and NeuMan~\cite{jiang2022neuman} for novel view synthesis at novel timesteps, where \method{} achieves better visual quality over NeuMan~\cite{jiang2022neuman} for the rendered human bodies and backgrounds in all scenes due to the superiority of our human model and background model. Our \method{} also achieves better foreground-background merging results, such as the feet regions of the Seattle scene and the Lab scene. This further demonstrates the effectiveness and flexibility of \method{} to model various types of dynamic human-(objects)-scenes.

\begin{table*}[h]
\centering{}%
\resizebox{\linewidth}{!}{

\begin{tabular}{c|ccc|cc|c|c|c|c}
\hline 
\multirow{2}{*}{Method} & \multicolumn{3}{c|}{Ours} & \multicolumn{2}{c|}{NeuMan~\cite{jiang2022neuman}} & \multirow{2}{*}{HyperNeRF~\cite{park2021hypernerf}} & \multirow{2}{*}{Nerfies~\cite{park2021nerfies}} & \multirow{2}{*}{D$^2$NeRF~\cite{wu2022d}} & \multirow{2}{*}{K-Planes~\cite{fridovich2023k}}\tabularnewline
 & 1st stage & 2nd stage & 3rd stage & 1st stage & 2nd stage &  &  &  & \tabularnewline
\hline 
No. of GPUs & 4 & 4 & 4 & 3 & 1 & 2 & 4 & 1 & 1\tabularnewline
Training time (hours) & 32 & 34 & 52 & 80 & 95 & 39 & 35 & 5.7 & 5.3\tabularnewline
\hline 
\end{tabular}
}
\caption{Training time comparison on the HOSNeRF dataset against baselines.}
\label{tab:traintime}
\vspace{-3mm}
\end{table*}

\noindent\textbf{Training Time Comparison on the \method{} Dataset.} Tab.~\ref{tab:traintime} presents the training time of all methods on our \method{} dataset. To ensure a fair comparison with the state-of-the-art (SOTA) approaches, we employ their highest configurations. Our three-stage training of \method{} requires a total of five days, whereas NeuMan's~\cite{jiang2022neuman} two-stage training demands over seven days. Due to the absence of distributed training support and the need for CPU computing, NeuMan's~\cite{jiang2022neuman} second stage training takes 95 hours. In contrast, although the training time for D$^2$NeRF~\cite{wu2022d} and K-Planes~\cite{fridovich2023k} is less than 6 hours, their performances are significantly inferior on our challenging dataset, as evidenced by Tab.~\ref{tab:hosnerf_comparison} and Fig.~\ref{fig:vis_result} of the main paper.

\noindent\textbf{Per-scene 360{\textdegree} Free-Viewpoint Renderings from Our \method{}.} We have also included a supplementary video to showcase the per-scene 360{\textdegree} free-viewpoint renderings from our \method{} on all six scenes of our \method{} dataset. The video highlights that our \method{} is capable of generating high-fidelity details that accurately resemble all scene components, including dynamic human bodies, objects, and backgrounds. Notably, our \method{} facilitates pausing the monocular video at any given point and rendering all scene details with high-fidelity from 360{\textdegree} viewpoints. To the best of our knowledge, our \method{} represents the first work to achieve 360{\textdegree} free-viewpoint high-fidelity novel view synthesis for dynamic scenes featuring human-environment interactions from a single video.

It should be emphasized that to enable 360{\textdegree} free-viewpoint rendering of the scene, the input single videos must contain 360{\textdegree} scene information (an example capture process is shown in the supplementary video); otherwise, artifacts will appear in unobserved areas that were not even seen in the input videos. However, our \method{} is not restricted by such capturing requirements, as demonstrated on the NeuMan dataset~\cite{jiang2022neuman}, which features small camera motions. It is worth noting that although the NeuMan dataset~\cite{jiang2022neuman} does not allow for 360{\textdegree} rendering since most scene regions are not captured, our \method{} still remains applicable.

\begin{figure*}
\begin{centering}
\includegraphics[width=1\linewidth]{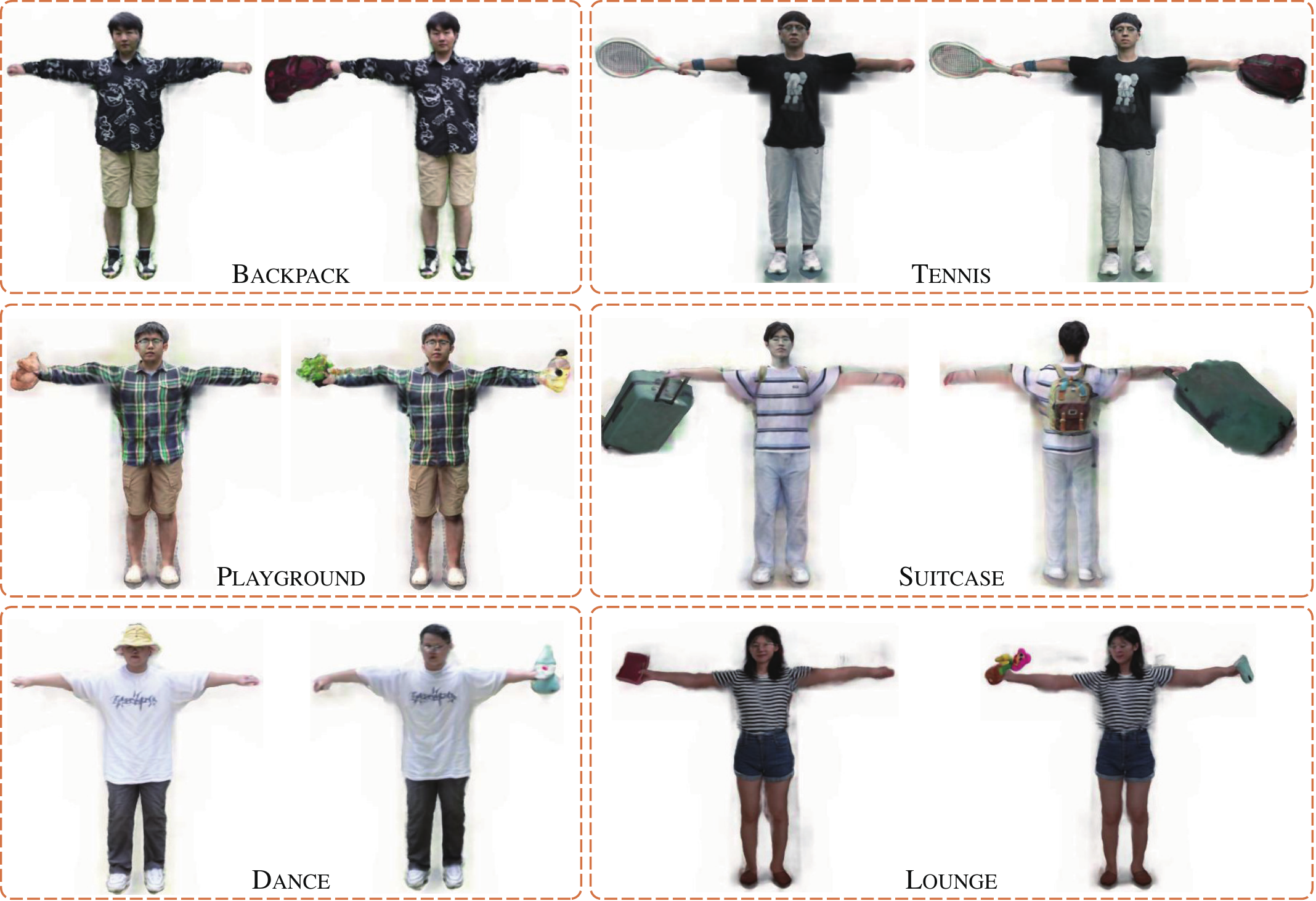}
\par\end{centering}
\caption{\label{fig:vis_canonical}Optimized state-conditional canonical spaces of \method{} on our \method{} dataset.}
\end{figure*}

\noindent\textbf{Optimized State-Conditional Canonical Spaces from Our \method{}.} Fig.~\ref{fig:vis_canonical} illustrates the state-conditional canonical spaces learned by our \method{} on the \method{} dataset. As shown in the figure, our proposed state-conditional dynamic human-object model can effectively represent different human-object states, and can reconstruct both the human bodies and objects with photorealistic details, enabling both 360{\textdegree} dynamic novel view synthesis and novel object / human pose manipulations. In addition, our complete \method{} is able to render clean human-object canonical spaces based on coarse human-object masks.

\end{document}